\documentclass[11pt,a4paper]{article}
\usepackage[hyperref]{emnlp2020}
\usepackage{multirow}
\usepackage{graphicx} 
\usepackage{float}
\usepackage{times}
\usepackage{latexsym}
\usepackage{CJKutf8}
\usepackage{amsmath}

\usepackage{color}
\usepackage{color}

\makeatletter
\g@addto@macro\normalsize{%
  \abovedisplayskip 4pt plus 2pt minus 3pt%
  \belowdisplayskip \abovedisplayskip
  \abovedisplayshortskip 4pt plus2pt  minus3pt%
  \belowdisplayshortskip 4pt plus2pt minus3pt%
}

\makeatother

\usepackage{microtype}

\aclfinalcopy 

\title{SlotRefine: A Fast Non-Autoregressive Model for\\ Joint Intent Detection and Slot Filling}
  
\author{Di Wu$^\mathsection$~~~~Liang Ding$^\dagger$~~~~Fan Lu$^\flat$~~~Jian Xie$^\ddagger$\\
~~$^\mathsection$Peking University~~~~~~~~$^\flat$University of Washington~~~~~~~~~~~~~~~$^\ddagger$Wuhan University~~~~~~~~~~~~~~\\
  {\tt inbath@163.com}~~~~~{\tt lufan0929@gmail.com}~~~~~{\tt xiejian1990@gmail.com}\\
  $^\dagger$UBTECH Sydney AI Centre, School of Computer Science\\ Faculty of Engineering,
  The University of Sydney\\
  {\tt ldin3097@uni.sydney.edu.au}
  }

\date{}

\begin{document}
\maketitle
\begin{abstract}
Slot filling and intent detection are two main tasks in spoken language understanding (SLU) system. In this paper, we propose a novel non-autoregressive model named SlotRefine for joint intent detection and slot filling. Besides, we design a novel two-pass iteration mechanism to handle the uncoordinated slots problem caused by conditional independence of non-autoregressive model. Experiments demonstrate that our model significantly outperforms previous models in slot filling task, while considerably speeding up the decoding (up to $\times$10.77). 
In-depth analyses show that 1) pretraining schemes could further enhance our model; 2) two-pass mechanism indeed remedy the uncoordinated slots.
\end{abstract}

\begin{figure*}[htp]
     \centering
     \includegraphics[width=1.0\textwidth]{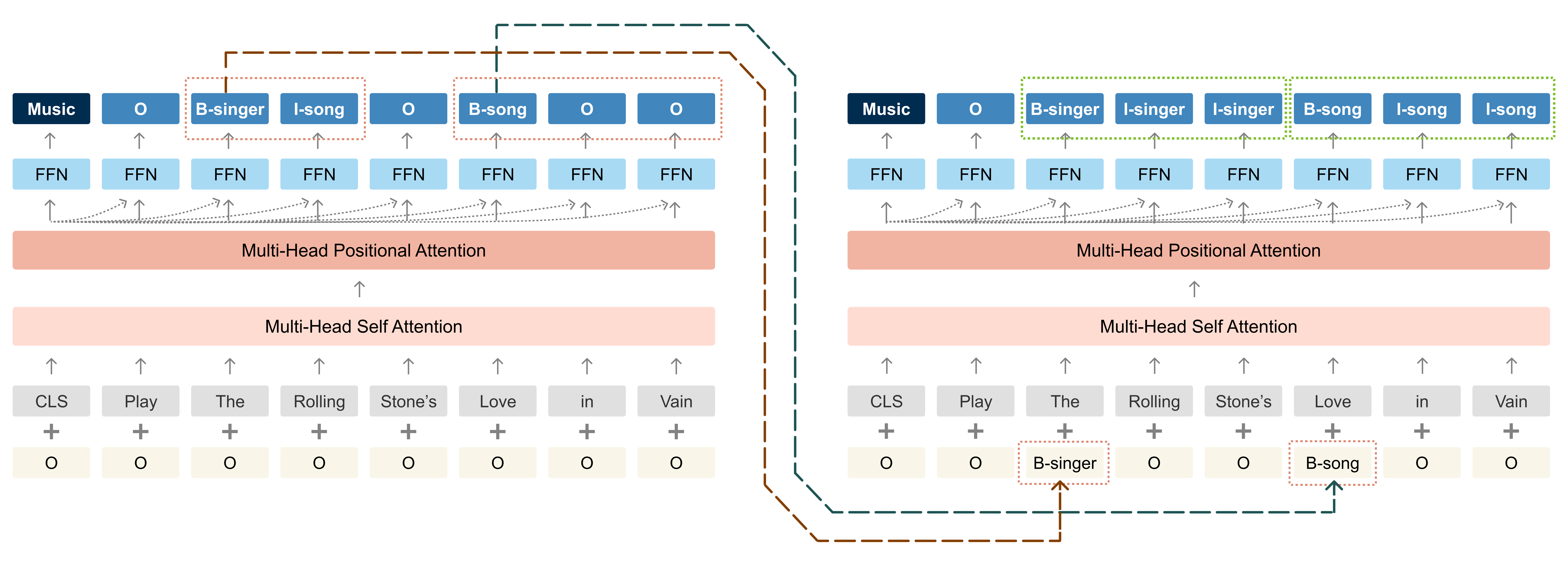}
     \caption{Illustration of SlotRefine, where the left and right part indicate the first and second iteration process respectively. In the first pass, wrong slot tagging results are predicted, as shown in the \textcolor[RGB]{217,117,137}{pink} dotted box in the figure, and the ``\textit{B-tags}'' (beginning tag of a slot) are feeded as additional information with utterance for second iteration. The slot results in the \textcolor[RGB]{116,186,41}{green} dotted box are refined results by second pass. Note that the initial tag embedding ``O'' added to each inputting position is designed for the two-pass mechanism(Sec.\S\ref{subsec:two-pass}).}
     \label{fig-1} 
\end{figure*}

\begin{figure}[tb]
     \centering
     \includegraphics[width=0.47\textwidth]{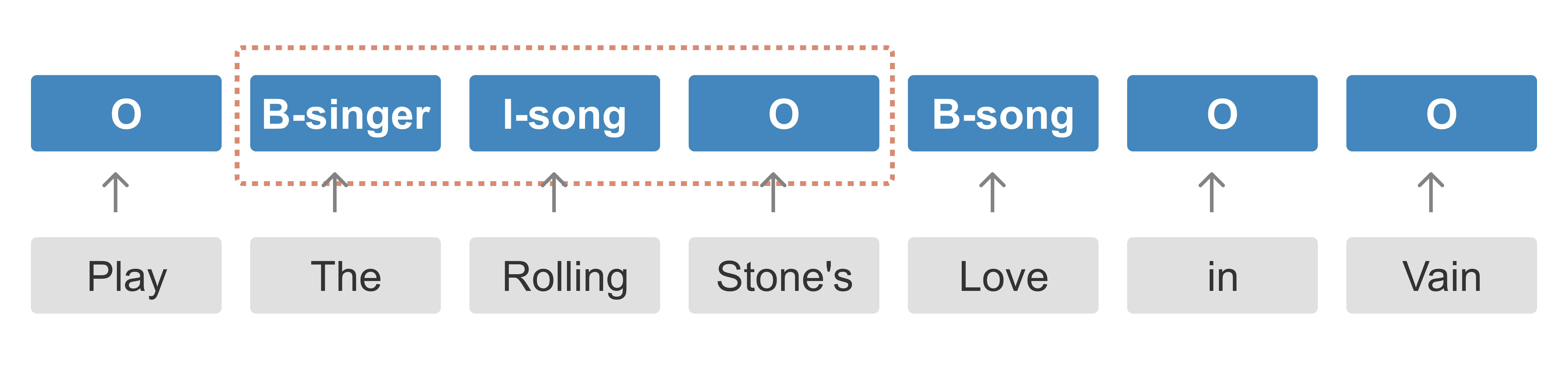}
     \caption{A example of uncoordinated slot tagging.}
     \label{fig-2}
\end{figure}

\section{Introduction}
Slot filling (SF) and intent detection (ID) play important roles in spoken language understanding, especially for task-oriented dialogue system. For example, for an utterance like ``\textit{Buy an air ticket from Beijing to Seattle}'', intent detection works on sentence-level to indicate the task is about purchasing an air ticket, while the slot filling focus on words-level to figure out the departure and destination of that ticket are ``\textit{Beijing}'' and ``\textit{Seattle}''.

In early studies, ID and SF were often modeled separately, where ID was modeled as a classification task, while SF was regarded as a sequence labeling task. Due to the correlation between these two tasks, training them jointly could enhance each other.  \newcite{zhang2016joint} propose a joint model using bidirectional gated recurrent unit to learn the representation at each time step. Meanwhile, a max-pooling layer is employed to capture the global features of a sentence for intent classification. \newcite{liu2016attention} cast the slot filling task as a tag generation problem and introduce a recurrent neural network based encoder-decoder framework 
with attention mechanism to model it, meanwhile using the encoded vector to predict intent. \newcite{goo2018slot} and \newcite{haihong2019novel} dig into the correlation between ID and SF deeper and modeled the relationship between them explicitly. ~\newcite{qin-etal-2019-stack} propagate the token-level intent results to the SF task, achieving significant performance improvement.

Briefly summarized, most of the previous works heavily rely on autoregressive approaches, \textit{e.g.,} RNN based model or seq2seq architecture, to capture the grammar structure in an utterance.
And conditional random field (CRF) is a popular auxiliary module for SF task as it considers the correlations between tags. 
Thus, several state-of-the-art works combine the autoregressive model and CRF to achieve the competitive performance, which therefore are set as our baseline methods.

However, for SF task, we argue that identifying token dependencies among slot chunk is enough, and it is unnecessary to model the entire sequence dependency in autoregressive fashion, which leads to redundant computation and inevitable high latency.

In this study, we cast these two tasks jointly as a non-autoregressive tag generation problem to get rid of unnecessary temporal dependencies. 
Particularly, a Transformer~\cite{vaswani2017attention} based architecture is adopted here to learn the representations of an utterance in both sentence and word level simultaneously (Sec.\S\ref{subsec:na-model}). The slots and intent labels are predicted independently and simultaneously, achieving better decoding efficiency.
We further introduce a two-pass refine mechanism (in Sec.\S\ref{subsec:two-pass}) to model boundary prediction of each slots explicitly, which also handle the uncoordinated slots problem (\textit{e.g.,} \textit{I-song} follows \textit{B-singer}) caused by conditional independence attribute. 

Experiments  on  two  commonly-cited datasets show that our approach is significantly and consistently superior to the existing models both in SF performance and efficiency (Sec.\S\ref{sec:exp}).
Our contributions are as follows: 
\begin{itemize}

\item We propose an fast non-autoregressive approach to model ID and SF tasks jointly, named SlotRefine\footnote{Our code is available:~\url{https://github.com/moore3930/SlotRefine}}, achieving the state-of-the-art on ATIS dataset.

\item We design a two-pass refine mechanism to handle uncoordinated slots problem. Our analyses confirm it is a better alternative than CRF in this task.

\item Our model infers nearly $\times$11 faster than existing models ($\times$13 for long sentences), indicating that our model has great potential for the industry and academia.

\end{itemize}

\section{Proposed Approaches}
\label{sec:our-approach}
 In this section, we first describe how we model slot filling and intent detection task jointly by an non-autoregressive model. And then we describe the details of the two-pass refine mechanism. 
The brief scheme of our model is shown in Figure~\ref{fig-1}, details can be found in the corresponding caption. Note that we follow the common practice~\cite{ramshaw-marcus-1995-text,zhang2016joint,haihong2019novel} to use ``Inside–outside–beginning (IOB)'' tagging format.

\begin{table*}[tp]
\centering
\scalebox{1}{
\begin{tabular}{l|lll|lll}
\hline\hline
\multirow{2}{*}{Model} & \multicolumn{3}{c|}{\textbf{ATIS Dataset}} & \multicolumn{3}{c}{\textbf{Snips Dataset}}
\cr\cline{2-7} & \textbf{Slot} & \textbf{Intent} & \textbf{Sent} & \textbf{Slot} & \textbf{Intent} & \textbf{Sent}\cr
\hline
Joint Seq \citep{hakkani2016multi} & 94.30 & 92.60 & 80.70& 87.30 & 96.90 & 73.20 \\
Atten.-Based \citep{liu2016attention} & 94.20 & 91.10 & 78.90 & 87.80 & 96.70 & 74.10\\
Sloted-Gated \citep{goo2018slot} & 95.42 & 95.41 & 83.73 & 89.27 & 96.86 & 76.43 \\
SF-ID (w/o CRF) \citep{haihong2019novel} & 95.50 & 96.58 & 86.00 & 90.46 & 97.00 & 78.37 \\
SF-ID (w/ CRF) \citep{haihong2019novel} & 95.80 & 97.09 & 86.90 & 92.23 & 97.29 & 80.43 \\
Stack-Propagation~\cite{qin-etal-2019-stack}&95.90&96.90&86.50&94.20&98.00&86.90\\
\hline
\textbf{Our Joint Model} (in Sec.\S\ref{subsec:na-model}) & 95.33 & 96.84 & 85.78 & 93.13 & 97.21 & 82.83 \\
\textbf{Our Joint Model +CRF} & 95.71 & 96.54 & 85.71 & 93.22 & 96.79 & 82.51 \\
\textbf{SlotRefine} & \textbf{96.22}$^\uparrow$ & \textbf{97.11}$^\uparrow$ & \textbf{86.96}$^\uparrow$ & \textbf{93.72} & 97.44 & 84.38 \\
\hline
\hline
\end{tabular}}
\caption{\label{table-1}Performance comparison on ATIS and Snips datasets. ``$\uparrow$''indicates significant difference ($p < 0.05$) with previous works. Model name written in bold refer to ours.}
\end{table*}

\subsection{Non-Autoregressively Joint Model}
\label{subsec:na-model}
We extend the original multi-head Transformer encoder in \citet{vaswani2017attention} to construct the model architecture of SlotRefine. Please refer to \citet{vaswani2017attention} for the details of Transformer. The main difference against the original Transformer is that we model the sequential information with relative position representations~\citep{shaw2018self}, instead of using absolute position encoding.

For a given utterance, a special token $\mathrm{CLS}$ is inserted to the first inputting position akin to the operation in BERT~\citep{devlin-etal-2019-bert}. Difference from that in BERT is the corresponding output vector is used for next sentence classification, we use it to predict the label of intent in SlotRefine. We denote the input sequence as $x = \left (x_{cls}, x_1, ..., x_l\right )$, where $l$ is the utterance length. Each word $x_i$ will be embedded into a $h$-dimention vector to perform the multi-head self-attention computation.
Then, the output of each model stack can be formulated as $H = \left (h_{cls}, h_1, ..., h_l\right )$. 

To jointly model the representations of ID and SF tasks, we directly concat~\footnote{We follow~\citep{goo2018slot} to fuse two representations with gating mechanism, but preliminary experiments show that simply concatenation performs best for our model structure.} the representations of $h_{cls}$ and $h_i$ before feed-forward computation, and then feed them into the softmax classifier. Specifically,
the intent detection and slot filling results are predicted as follows, respectively:
\begin{equation}
\begin{aligned}
        y^{i} &= \mathrm{softmax} \left ( W^{i} \cdot h_{cls} + b^{i} \right )\\
    y_{i}^{s} &= \mathrm{softmax} \left ( W^{s} \cdot\left [  h_{cls}, h_{i}\right ]  + b^{s} \right )
\end{aligned}
\end{equation}
where $y^{i}$ and $y_{i}^{s}$ denote intent label of the utterance and slot label for each token $i$, respectively. $\left [  h_{cls}, h_{i}\right ]$ is the concated vector. $W$ and $b$ are corresponding trainable parameters. 

The objective of our joint model can be formulated as:
\begin{equation}
    p\left ( y^{i}, y^{s} | x \right ) = p\left ( y^{i} | x \right ) \cdot \prod_{t}^{l} p\left ( y_{t}^{s} | x, y^{i} \right )
\end{equation}
The learning objective is to maximize the conditional probability $p\left ( y^{i}, y^{s} | x \right )$, which is optimized via minimizing its cross-entropy loss. Unlike autoregressive methods, the likelihood of each slot in our approach can be optimized in parallel.

\subsection{Two-pass Refine Mechanism}
\label{subsec:two-pass}
Due to the conditional independence between slot labels, it is difficult for our proposed non-autoregressive model to capture the sequential dependency information among each slot chunk, thus leading to some uncoordinated slot labels. We name this problem as \textbf{uncoordinated slots problem}. Take the false tagging in Figure~\ref{fig-2} for example, slot label ``\textit{I-song}'' uncoordinately follows ``\textit{B-singer}'', which does not satisfy the Inside-Outside-Beginning tagging format.

To address this problem, we introduce a two-pass refine mechanism. As depicted in the Figure~\ref{fig-1}, in addition to each token embedding in the utterance, we also element-wisely add the slot tag embedding into the model. In the first pass, the initial slot tags are all setting to ``O'', while in the second pass, the ``\textit{B-tags}'' predicted in the first pass is used as the corresponding slot tag input. These two iterations share the model and optimization goal, thus brings no extra parameters. 

Intuitively, in doing so, the model generates a draft in the first pass and tries to find the beginning of each slot chunk. In the second pass, by propagating the utterance again with the predicted ``\textit{B-tags}'', the model is forced to learn how many identical ``\textit{I-tags}'' follow them. Through this process, the slot labels predicted becomes more consistent, and the boundaries are more accurately identified.
From a more general perspective, we can view this two-pass process as a trade-off between autoregression and non-autoregression, where the complete markov chain process can be simplified as follow: 
\begin{equation}
    \begin{split}
        p\left ( y^{i}, y^{s} | x \right ) &= p\left ( y^{i} | x \right ) \cdot p\left ( \tilde{y}^{s} | y^{i}, x \right ) \\
        & \cdot p\left ( y^{s} | \tilde{y}^{s}, y^{i}, x \right )
    \end{split}
\end{equation}
where $\tilde{y}^{s}$ is the tagging results from the first pass.

Two-pass refine mechanism is similar to the multi-round iterative mechanism in non-autoregressive machine translation~\cite{lee2018deterministic, gu2017non,ding2020localness,kasai2020parallel}, such as Mask-predict~\citep{ghazvininejad2019mask}. However, we argue that our method is more suitable in this task. The label dependency of the tagging task (e.g., slot filling) is simple, where we only need to ensure the tagging labels of a slot are consistent from the beginning to the end. Therefore, two iterations to force the model to focus on the slot boundaries is enough in our task, intuitively. Mask-Predict can alleviate the problem caused by conditional independence too. However, it’s designed for a more complex goal, and it usually introduce more iterations (e.g., 10 iters) to achieve competitive performance, which largely reduces the inference speed.

\section{Experiment} 
\label{sec:exp}
\paragraph{Datasets}

We choose two widely-used datasets: ATIS (Airline Travel Information Systems,\citet{tur2010left}) and Snips (collected by Snips personal voice assistant,\citet{coucke2018snips}). Compared with ATIS, the Snips dataset is more complex due to its large vocabulary size, cross-domain intents and more out-of-vocabulary words.

\paragraph{Metrics} 
Three evaluation metrics are used in our experiments. F1-score and accuracy are applied for slot filling and intent detection task, respectively. Besides, we use sentence accuracy to indicate proportion of utterance in the corpus whose slots and intent are both correctly-predicted.

\paragraph{Setup}

All embeddings are initialized with xavier method~\cite{glorot2010understanding}. The batch size is set to 32 and learning rate is 0.001. we set number of Transformer layers, attention heads and hidden sizes to \{2,8,64\} and \{4,16,96\} for ATIS and Snips datasets. In addition, we report the results of previous studies ~\citep{hakkani2016multi,liu2016attention,goo2018slot,haihong2019novel,qin-etal-2019-stack} and conduct speed evaluation based on their open-source codes.

\paragraph{Main Results}
\label{subsec:mainresults}
Table~\ref{table-1} summarizes the model performance on ATIS and snips corpus. It can be seen that SlotRefine consistently outperforms other baselines in all three metrics. 
Compared with our basic non-autoregressive joint model in Section\S~\ref{subsec:na-model}, SlotRefine achieve +1.18 and +1.55 sentence-level accuracy improvements for ATIS and Snips, respectively. It is worthy noting that our SlotRefine significantly improves the slot filling task (F1-score$\uparrow$). we attribute the improvement to that our two-pass mechanism successfully makes the model learn better slot boundaries.

\begin{table}[tp]
\centering
\scalebox{0.95}{
\begin{tabular}{lrr}
\hline\hline
\textbf{Model} & \textbf{Latency} & \textbf{Speedup} \cr
\hline
Sloted-Gated & 11.31ms & $1.41\times$ \\
SF-ID (with CRF) & 13.03ms & $1.22\times$ \\
Stack-Propagation & 15.94ms & $1.00\times$\\
\hline
\textbf{Our Joint Model} & 1.48ms & $10.77\times$ \\
\textbf{Our Joint Model +CRF} & 8.32ms & $1.92\times$ \\
\textbf{SlotRefine} & 3.02ms & $4.31\times$ \\
\hline
\hline
\end{tabular}}
\caption{\label{table-2} ``Latency'' is the average time to decode an utterance without minibatching. ``Speedup'' is compared against existing SOTA model~\cite{haihong2019novel}.}
\end{table}

\paragraph{Speedup}
\label{subsec:speed}
As each slot tagging result can be calculated in parallel with our approach, inference latency can be significantly reduced.
As shown in Table~\ref{table-2}, on ATIS test set, our non-autoregressive model could achieve $\times$8.80 speedup compared with the existing state-of-the-art model~\cite{haihong2019novel}. And after introducing two-pass mechanism (SlotRefine), our model still achieves competitive inference speedup ($\times$4.31). Our decoding is conducted with a single Tesla P40 GPU. It is worth noting that for long sentences (Length$\geq$12), the speedup achieves $\times$13 (not reported in table).

\paragraph{Two-Pass Mechanism v.s. CRF}
In SF task, CRF is usually used to learn the dependence of slot labels. Two most important dependence rules CRF learned can be summarized as \textit{tag O can only be followed by O or B } and \textit{tag B-* can only be followed by same-type label I-* or O}, 
which can be perfectly addressed with our proposed two-pass mechanism. 
Experiments about +CRF can be found in Table~\ref{table-1}\&\ref{table-2} (``Our Joint Model +CRF''), we can see that two-pass mechanism equipped SlotRefine outperforms +CRF by averagely +0.89, meanwhile preserving $\times$2.8 speedup, demonstrating that two-pass mechanism can be a better substitute for CRF in this task for better performance and efficiency.

\begin{figure}[tp]
     \centering
     \includegraphics[width=0.45\textwidth]{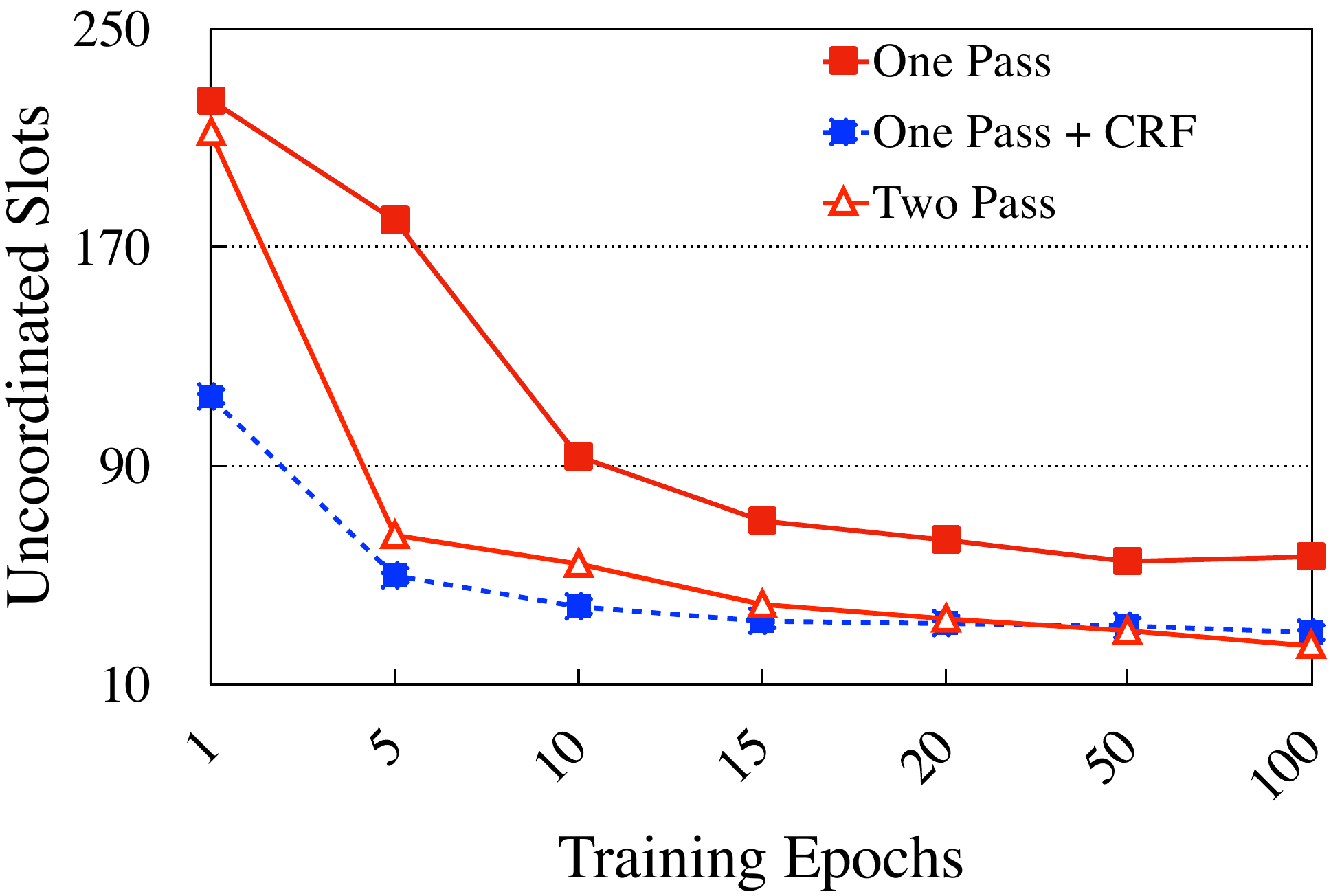}
     \caption{The number of uncoordinated slots of our joint model (One Pass), joint model with CRF (One Pass+CRF) and SlotRefine (Two Pass) during training.}
     \label{fig-3}
\end{figure}

\begin{table*}[htb]
\centering
\scalebox{0.99}{
\begin{tabular}{l|ccc|ccc}
\hline\hline
\multirow{2}{*}{Model} & \multicolumn{3}{c|}{\textbf{ATIS Dataset}} & \multicolumn{3}{c}{\textbf{Snips Dataset}}
\cr\cline{2-7} & \textbf{Slot} & \textbf{Intent} & \textbf{Sent} & \textbf{Slot} & \textbf{Intent} & \textbf{Sent}\cr
\hline
Joint Model & 95.33 & 96.84 & 85.78 & 93.31 & 97.21 & 82.83 \\
Joint Model with CRF & 95.71 & 96.54 & 85.71 & 93.22 & 96.79 & 82.51 \\
SlotRefine & 96.22 & 97.11 & 86.96 & 93.72 & 97.44 & 84.38 \\
\hline
\textbf{SlotRefine} with GloVe &\bf 96.24 & 97.35 & 87.57 & 96.33 & 98.36 & 91.06 \\
\textbf{SlotRefine} with BERT & 96.16 &\bf 97.74 &\bf 88.64 &\bf 97.05 &\bf 99.04 &\bf 92.96 \\ 
\hline
\multicolumn{7}{c}{previous work with pretraining}\\
\hline
\textbf{BERT-Joint} \citep{chen2019bert} & 96.10 & 97.50 & 88.20 &  97.00 & 98.60 & 92.80 \\
\textbf{Stack-Propagation} with BERT~\cite{qin-etal-2019-stack}&96.10&97.50&88.60&97.00&99.00&92.90\\
\hline
\hline
\end{tabular}}
\caption{\label{table-3}Performance comparison between SlotRefine with GloVe initialzation and Bert based model on ATIS and Snips datasets.}
\end{table*}

\paragraph{Remedy Uncoordinated Slots in Training}
We visualize the number decrease of uncoordinated slots of the training process on ATIS dataset. As depicted in Figure~\ref{fig-3}, uncoordinated errors of both ``One-Pass'' and ``Two-Pass'' models decrease with training goes. Notably, the uncoordinated slots number of Two-Pass model drops significantly faster than the One-Pass model, achieving better convergence than +CRF after 50 epochs. This indicates that our proposed two-pass mechanism indeed remedy the uncoordinated slots problem, making the slot filling more accurate.


\paragraph{SlotRefine with Pretraining}
Recently, there are also some works based on large scale pretraining model BERT~\citep{chen2019bert}, where billions of external corpus are used and tremendous of model parameters are introduced.
The number of parameters of BERT is many orders of magnitude more than ours, thus it is unfair to compare performance of SlotRefine with them directly. To highlight the effectiveness of SlotRefine, we conduct experiments with two pretraining schemes, GloVe\footnote{https://github.com/stanfordnlp/GloVe} and BERT\footnote{https://github.com/huggingface/transformers}, to compare with them. We find that both GloVe and BERT could further enhance the SlotRefine, and it worth noting that ``SlotRefine w/ BERT'' outperforms existing pretraining based models. The detailed comparison can be found in Table~\ref{table-3}.

For the pre-training scheme of BERT, we follow the setting in~\citet{chen2019bert} and equip two-pass mechanism in the fine-tune stage, where CLS token is used for intent detection. And for the pre-training scheme of GloVe, we fix and compress the pretrained word vectors into the same dimension of the input hidden size in SlotRefine by a dense network. It is worth noting that through such simple pre-training method, SlotRefine can achieve a results very close to the method implemented by BERT. We guess that the benefits of the pre-training methods on this task mainly come from alleviating the Out-of-Vocabulary (OOV) problem. One piece of evidence is, for Snips whose test set has a large number of OOV words, benefits through pre-training are very obvious. However, for the ATIS whose test set has few OOV words, only a small sentence accuracy gain, 0.61 and 1.68 for GloVe and Bert respectivly, is obtained after using the pre-training method.

\section{Conclusion}
In this paper, we first reveal an \textit{uncoordinated slots problem} for a classical language understanding task, i.e., slot filling. To address this problem, we present a novel non-autoregressive joint model for slot filling and intent detection with two-pass refine mechanism (non-autoregressive refiner), which significantly improves the performance while substantially speeding up the decoding. Further analyses show that our proposed non-autoregressive refiner has great potential to replace CRF in at least slot filling task.


In the future, we plan to extend our non-autoregressive refiner to other Natural Language Understanding (NLU) tasks, e.g., named entity recognition~\cite{tjong-kim-sang-de-meulder-2003-introduction}, semantic role labeling~\cite{he-etal-2018-jointly}, and Natural Language Generation (NLG) tasks, e.g., machine translation~\cite{vaswani2017attention}, summarization~\cite{liu2019text}.

\section*{Acknowledgments}
We thank the anonymous reviewers for their helpful suggestions. We gratefully acknowledge the support of DuerOS department, Baidu company. Any opinions, findings, and conclusions or recommendations expressed in this material are those of the authors and do not necessarily reflect the view of Baidu company.


\bibliography{emnlp2020}

\begin{thebibliography}{22}
\expandafter\ifx\csname natexlab\endcsname\relax\def\natexlab#1{#1}\fi

\bibitem[{Chen et~al.(2019)Chen, Zhuo, and Wang}]{chen2019bert}
Qian Chen, Zhu Zhuo, and Wen Wang. 2019.
\newblock Bert for joint intent classification and slot filling.
\newblock In \emph{arXiv}.

\bibitem[{Coucke et~al.(2018)Coucke, Saade, Ball, Bluche, Caulier, Leroy,
  Doumouro, Gisselbrecht, Caltagirone, Lavril et~al.}]{coucke2018snips}
Alice Coucke, Alaa Saade, Adrien Ball, Th{\'e}odore Bluche, Alexandre Caulier,
  David Leroy, Cl{\'e}ment Doumouro, Thibault Gisselbrecht, Francesco
  Caltagirone, Thibaut Lavril, et~al. 2018.
\newblock Snips voice platform: an embedded spoken language understanding
  system for private-by-design voice interfaces.
\newblock In \emph{arXiv}.

\bibitem[{Devlin et~al.(2019)Devlin, Chang, Lee, and
  Toutanova}]{devlin-etal-2019-bert}
Jacob Devlin, Ming-Wei Chang, Kenton Lee, and Kristina Toutanova. 2019.
\newblock {BERT}: Pre-training of deep bidirectional transformers for language
  understanding.
\newblock In \emph{NAACL}.

\bibitem[{Ding et~al.(2020)Ding, Wang, Wu, Tao, and Tu}]{ding2020localness}
Liang Ding, Longyue Wang, Di~Wu, Dacheng Tao, and Zhaopeng Tu. 2020.
\newblock Context-aware cross-attention for non-autoregressive translation.
\newblock In \emph{COLING}.

\bibitem[{Ghazvininejad et~al.(2019)Ghazvininejad, Levy, Liu, and
  Zettlemoyer}]{ghazvininejad2019mask}
Marjan Ghazvininejad, Omer Levy, Yinhan Liu, and Luke Zettlemoyer. 2019.
\newblock Mask-predict: Parallel decoding of conditional masked language
  models.
\newblock In \emph{EMNLP}.

\bibitem[{Glorot and Bengio(2010)}]{glorot2010understanding}
Xavier Glorot and Yoshua Bengio. 2010.
\newblock Understanding the difficulty of training deep feedforward neural
  networks.
\newblock In \emph{ICML}.

\bibitem[{Goo et~al.(2018)Goo, Gao, Hsu, Huo, Chen, Hsu, and
  Chen}]{goo2018slot}
Chih-Wen Goo, Guang Gao, Yun-Kai Hsu, Chih-Li Huo, Tsung-Chieh Chen, Keng-Wei
  Hsu, and Yun-Nung Chen. 2018.
\newblock Slot-gated modeling for joint slot filling and intent prediction.
\newblock In \emph{NAACL}.

\bibitem[{Gu et~al.(2018)Gu, Bradbury, Xiong, Li, and Socher}]{gu2017non}
Jiatao Gu, James Bradbury, Caiming Xiong, Victor~OK Li, and Richard Socher.
  2018.
\newblock Non-autoregressive neural machine translation.
\newblock In \emph{ICLR}.

\bibitem[{Haihong et~al.(2019)Haihong, Niu, Chen, and Song}]{haihong2019novel}
E~Haihong, Peiqing Niu, Zhongfu Chen, and Meina Song. 2019.
\newblock A novel bi-directional interrelated model for joint intent detection
  and slot filling.
\newblock In \emph{ACL}.

\bibitem[{Hakkani-T{\"u}r et~al.(2016)Hakkani-T{\"u}r, T{\"u}r, Celikyilmaz,
  Chen, Gao, Deng, and Wang}]{hakkani2016multi}
Dilek Hakkani-T{\"u}r, G{\"o}khan T{\"u}r, Asli Celikyilmaz, Yun-Nung Chen,
  Jianfeng Gao, Li~Deng, and Ye-Yi Wang. 2016.
\newblock Multi-domain joint semantic frame parsing using bi-directional
  rnn-lstm.
\newblock In \emph{Interspeech}.

\bibitem[{He et~al.(2018)He, Lee, Levy, and Zettlemoyer}]{he-etal-2018-jointly}
Luheng He, Kenton Lee, Omer Levy, and Luke Zettlemoyer. 2018.
\newblock Jointly predicting predicates and arguments in neural semantic role
  labeling.
\newblock In \emph{ACL}.

\bibitem[{Kasai et~al.(2020)Kasai, Cross, Ghazvininejad, and
  Gu}]{kasai2020parallel}
Jungo Kasai, James Cross, Marjan Ghazvininejad, and Jiatao Gu. 2020.
\newblock Parallel machine translation with disentangled context transformer.
\newblock In \emph{ICML}.

\bibitem[{Lee et~al.(2018)Lee, Mansimov, and Cho}]{lee2018deterministic}
Jason Lee, Elman Mansimov, and Kyunghyun Cho. 2018.
\newblock Deterministic non-autoregressive neural sequence modeling by
  iterative refinement.
\newblock In \emph{EMNLP}.

\bibitem[{Liu and Lane(2016)}]{liu2016attention}
Bing Liu and Ian Lane. 2016.
\newblock Attention-based recurrent neural network models for joint intent
  detection and slot filling.
\newblock \emph{Interspeech}.

\bibitem[{Liu and Lapata(2019)}]{liu2019text}
Yang Liu and Mirella Lapata. 2019.
\newblock Text summarization with pretrained encoders.
\newblock In \emph{EMNLP}.

\bibitem[{Qin et~al.(2019)Qin, Che, Li, Wen, and Liu}]{qin-etal-2019-stack}
Libo Qin, Wanxiang Che, Yangming Li, Haoyang Wen, and Ting Liu. 2019.
\newblock A stack-propagation framework with token-level intent detection for
  spoken language understanding.
\newblock In \emph{EMNLP}.

\bibitem[{Ramshaw and Marcus(1995)}]{ramshaw-marcus-1995-text}
Lance Ramshaw and Mitch Marcus. 1995.
\newblock Text chunking using transformation-based learning.
\newblock In \emph{Third Workshop on Very Large Corpora}.

\bibitem[{Shaw et~al.(2018)Shaw, Uszkoreit, and Vaswani}]{shaw2018self}
Peter Shaw, Jakob Uszkoreit, and Ashish Vaswani. 2018.
\newblock Self-attention with relative position representations.
\newblock In \emph{NAACL}.

\bibitem[{Tjong Kim~Sang and
  De~Meulder(2003)}]{tjong-kim-sang-de-meulder-2003-introduction}
Erik~F. Tjong Kim~Sang and Fien De~Meulder. 2003.
\newblock Introduction to the {C}o{NLL}-2003 shared task: Language-independent
  named entity recognition.
\newblock In \emph{NAACL}.

\bibitem[{Tur et~al.(2010)Tur, Hakkani-T{\"u}r, and Heck}]{tur2010left}
Gokhan Tur, Dilek Hakkani-T{\"u}r, and Larry Heck. 2010.
\newblock What is left to be understood in atis?
\newblock In \emph{2010 IEEE Spoken Language Technology Workshop}. IEEE.

\bibitem[{Vaswani et~al.(2017)Vaswani, Shazeer, Parmar, Uszkoreit, Jones,
  Gomez, Kaiser, and Polosukhin}]{vaswani2017attention}
Ashish Vaswani, Noam Shazeer, Niki Parmar, Jakob Uszkoreit, Llion Jones,
  Aidan~N Gomez, {\L}ukasz Kaiser, and Illia Polosukhin. 2017.
\newblock Attention is all you need.
\newblock In \emph{NIPS}.

\bibitem[{Zhang and Wang(2016)}]{zhang2016joint}
Xiaodong Zhang and Houfeng Wang. 2016.
\newblock A joint model of intent determination and slot filling for spoken
  language understanding.
\newblock In \emph{IJCAI}.

\end{thebibliography}
\bibliographystyle{acl_natbib}

\appendix
\newpage

\end{document}